\documentclass[12pt]{article}

\usepackage{newtxtext,newtxmath}

\usepackage{graphicx}
\usepackage{wrapfig}
\usepackage{hyperref}

\usepackage{siunitx}

\usepackage[letterpaper,margin=1in]{geometry}

\renewenvironment{abstract}
	{\quotation}
	{\endquotation}

\date{}

\makeatletter
\renewcommand{\fnum@figure}{\textbf{Figure \thefigure}}
\renewcommand{\fnum@table}{\textbf{Table \thetable}}
\makeatother

\usepackage{scicite}

\usepackage{url}
\usepackage{placeins}

\def\scititle{
    Head-to-Head autonomous racing at the limits of handling in the A2RL challenge
}
\title{\bfseries \boldmath \scititle}

\author{
	Simon Hoffmann$^{1\ast\dagger}$,
	Simon Sagmeister$^{1\ast\dagger}$,
        Tobias Betz$^{1}$,
        Joscha Bongard$^{2}$,\and
        Sascha Büttner$^{1}$,
        Dominic Ebner$^{1}$,
        Daniel Esser$^{1}$,
        Georg Jank$^{2}$,\and
        Sven Goblirsch$^{1}$,
        Alexander Langmann$^{3}$,
        Maximilian Leitenstern$^{1}$,\and
        Levent Ögretmen$^{2}$,
        Phillip Pitschi$^{2}$,
        Ann-Kathrin Schwehn$^{3}$, \and
        Cornelius Schröder$^{1}$, 
        Marcel Weinmann$^{1}$,
        Frederik Werner$^{3}$,\and
         Boris Lohmann$^{2}$, Johannes Betz$^{3}$, Markus Lienkamp$^{1}$\and
	\small$^{1}$Institute of Automotive Technology, TUM, 85748, Garching, Deutschland.\and
	\small$^{2}$Chair of Automatic Control, TUM, 85748, Garching, Deutschland.\and
        \small$^{3}$Professorship Autonomous Vehicle Systems, TUM, 85748, Garching, Deutschland.\and
	\small$^\ast$Corresponding authors. Email: \{firstname\}.\{lastname\}@tum.de\and
	\small$^\dagger$These authors contributed equally to this work.
}

\usepackage{tikz}
\usepackage{pgfplots}
\usepgfplotslibrary{statistics}
\usepgfplotslibrary{fillbetween}
\usepgfplotslibrary{groupplots}
\usepackage{comment}
\usepackage{subcaption}
\usepackage{graphicx}

\usepackage[inkscapelatex=false, inkscapepath=./build/svg-inkscape]{svg}

\pgfplotsset{compat=1.18}

\definecolor{tumblue}{RGB}{0,101,189}
\definecolor{darkbluetum}{RGB}{0,82,147}
\definecolor{darkerbluetum}{RGB}{0,51,89}
\definecolor{orangetum}{RGB}{227,114,34}
\definecolor{greentum}{RGB}{162,173,0}
\definecolor{lightbluetum}{RGB}{152,198,234}
\definecolor{pastelbluetum}{RGB}{100,160,200}
\definecolor{pinkdark}{RGB}{155, 70, 141}
\definecolor{yellow}{RGB}{254, 215, 2}

\definecolor{TUMBlue}{RGB}{0,101,189}%
\definecolor{TUMWhite}{RGB}{255,255,255}%
\definecolor{TUMBlack}{RGB}{0,0,0}%
\definecolor{TUMBlue1}{RGB}{0,51,89}%
\definecolor{TUMBlue2}{RGB}{0,82,147}%
\definecolor{TUMGray1}{RGB}{51,51,51}%
\definecolor{TUMGray2}{RGB}{127,127,127}%
\definecolor{TUMGray3}{RGB}{204,204,204}%
\definecolor{TUMBlue3}{RGB}{100,160,200}%
\definecolor{TUMBlue4}{RGB}{152,198,234}%
\definecolor{TUMIvory}{RGB}{218,215,203}%
\definecolor{TUMOrange}{RGB}{227,114,34}%
\definecolor{TUMGreen}{RGB}{162,173,0}%

\newcommand{\ros}{ROS~2}

\definecolor{darkgray}{RGB}{32, 32, 32}

\pgfplotsset{
    compat=1.18,
    standard-plot/.style={
            enlarge x limits=0,
            grid=major,
            grid style={dotted},
        },
    line-plot/.style={
            standard-plot,
            width=\columnwidth,
            height=0.5\columnwidth,
            every tick label/.append style={font=\footnotesize},
            label style={font=\footnotesize},
        },
    xy-plot/.style={
            width=\columnwidth,
            height=0.5\columnwidth,
            standard-plot,
        },
    group-plot/.style={
            width=\textwidth,
            every tick label/.append style={font=\footnotesize}
        },
    group-plot-5/.style={
            group-plot,
            group style={group size=1 by 5,
                    xlabels at=edge bottom,
                    xticklabels at=edge bottom,
                    vertical sep=0.15cm,
                },
        },
    first-group-plot/.style={
            standard-plot,
            try min ticks=5,
            enlarge x limits=0,
            enlarge y limits=0.1,
            xmin=0,
            legend style={
                    at={(0.99, 0.5)},anchor=east,
                    nodes={scale=0.8, transform shape}
                },
            y label style={
                },
            label style={font=\footnotesize},
        },
    middle-group-plot/.style={
            standard-plot,
            enlarge x limits=0,
            enlarge y limits=0.1,
            xmin=0,
            try min ticks=5,
            legend style={
                    at={(0.99, 0.5)},anchor=east,
                    nodes={scale=0.8, transform shape}
                },
            y label style={
                },
            label style={font=\footnotesize},
        },
    last-group-plot/.style={
            standard-plot,
            xlabel=$t$ / \si{\second},
            enlarge x limits=0,
            enlarge y limits=0.1,
            try min ticks=5,
            xmin=0,
            legend style={
                    at={(0.99, 0.5)},anchor=east,
                    nodes={scale=0.8, transform shape}
                },
            y label style={
                },
            label style={font=\footnotesize},
        },
    generic-linestyle/.style={thick, mark size=1.0pt},
    reference/.style={generic-linestyle, TUMBlack},
    line-1/.style={generic-linestyle, TUMBlue2},
    line-2/.style={generic-linestyle, black},
    line-3/.style={generic-linestyle, TUMGreen},
    line-target/.style={generic-linestyle, TUMOrange},
    line-actual/.style={generic-linestyle, TUMBlue2},
}

\begin{document}

\maketitle
\begin{abstract} \bfseries \boldmath

Autonomous racing presents a complex challenge involving multi-agent interactions between vehicles operating at the limit of performance and dynamics. As such, it provides a valuable research and testing environment for advancing autonomous driving technology and improving road safety.
This article presents the algorithms and deployment strategies developed by the TUM Autonomous Motorsport team for the inaugural Abu Dhabi Autonomous Racing League (A2RL). We showcase how our software emulates human driving behavior, pushing the limits of vehicle handling and multi-vehicle interactions to win the A2RL.
Finally, we highlight the key enablers of our success and share our most significant learnings. %

\end{abstract}

\section{Introduction} %

\subsection{Motivation}
Autonomous racing has become increasingly popular over the last few years~\cite{Betz2022, wurman2022outracing}. This is due to competitions that provide events and platforms for evaluating Autonomous Driving~(AD) software in a competitive racing environment. Autonomous racing is a domain where overly conservative policies are penalized, demanding a balance between safety and performance. To compete at the highest level, algorithms must operate vehicles at the handling limits while strategically interacting with non-cooperative agents.
Furthermore, the algorithms must deal with highly adversarial environments, where even the slightest errors in perception, localization, planning, or control can have severe consequences. Minor driving mistakes can result in costly crashes, loss of competition, or wasted testing time.
If one had to choose a human driver for a critical driving situation, one would likely select a Formula1 driver with superior ability to understand the vehicle's behavior and make split-second decisions under extreme conditions. We aim to develop the equivalent of a Formula1 driver for AD software.

\subsection{The Abu Dhabi Autonomous Racing League}

In April 2024, eight teams worldwide participated in the first Abu Dhabi Autonomous Racing League (A2RL), a  \num{2.25} million dollar robotics competition. The A2RL was held at the Yas Marina Circuit in Abu Dhabi, a venue known for hosting Formüula1 races. The competing teams were Code19 Racing, Constructor University, Fly Eagle, HUMDA, Kinetiz, PoliMOVE, Unimore, and TUM Autonomous Motorsport.

This competition aimed to perform the first autonomous multi-vehicle race on a Formula1 track, delivering a human-like racing experience. The team's challenge was to develop the software while the organizers provided the race cars. 
Each session started on cold tires, and teams were forbidden to interact with the car during official sessions. Therefore, the software had to be able to handle the vehicle behavior during tire warm-up by itself. 

The competition consisted of three individual challenges: a time trial, an attack-and-defend, and the grand final with a four-vehicle race. In the time trial, each team had three sessions with two full laps to set the fastest time. 
In the attack-and-defend competition, two teams competed in two runs, switching the starting order for the second round. The goal was to overtake the leading vehicle and finish the run first. 
In the grand final, the four fastest teams from the time trial r aced head-to-head against each other after a rolling start. Whichever car finishes first after seven laps of free racing is the competition's winner. 
\begin{figure}
    \centering
    \begin{subfigure}[t]{0.485\textwidth}
        \centering
        \includegraphics[width=1\linewidth]{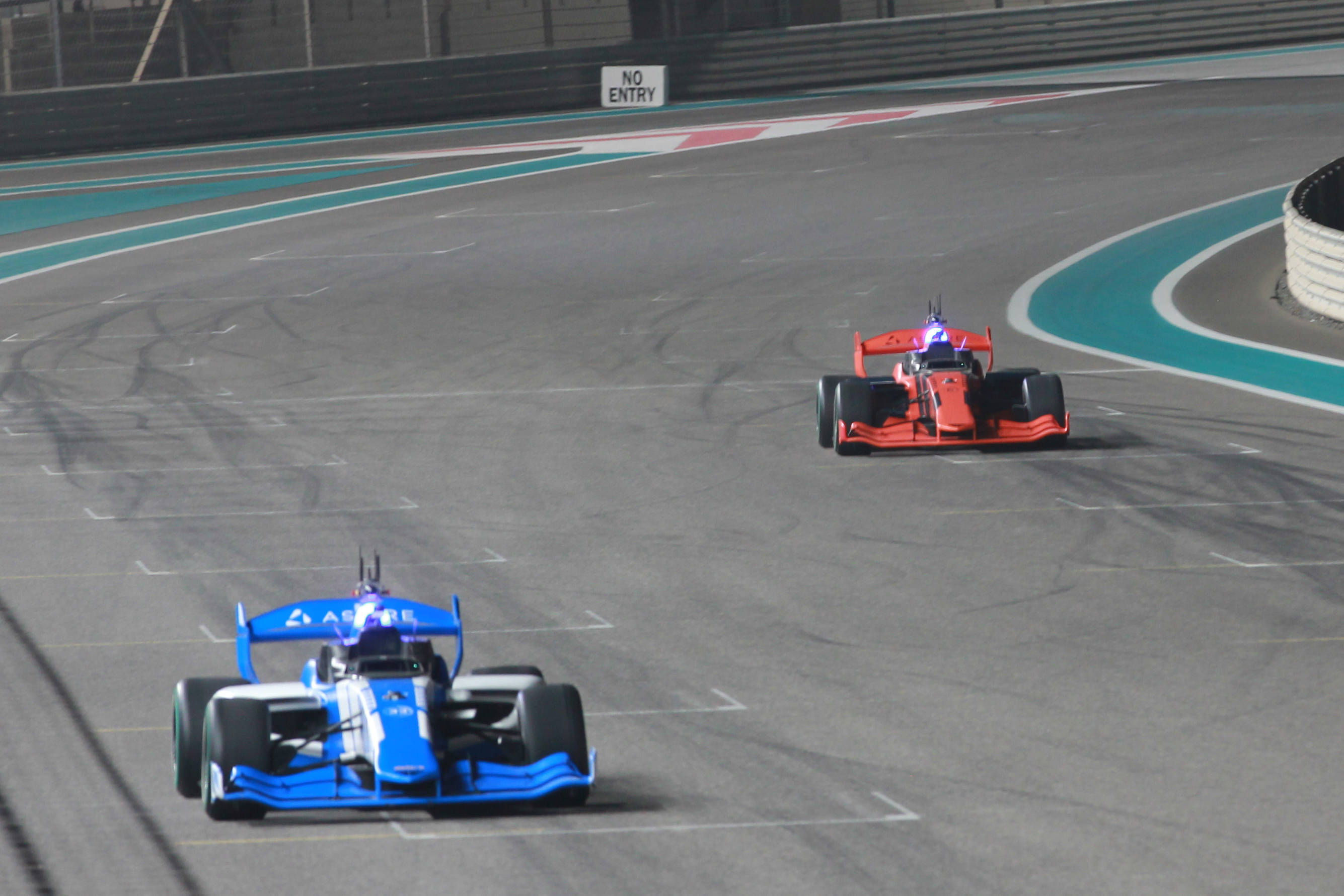}
        \caption{Two EAV24 race cars on the Yas Marina Circuit.}
    \end{subfigure}%
    \hfill
    \begin{subfigure}[t]{0.485\textwidth}
        \centering
        \includegraphics{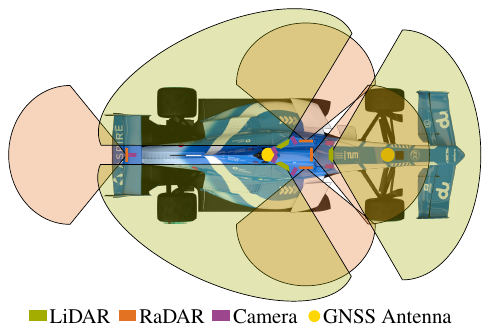}
        \caption{Sensor setup of the EAV24.}
    \end{subfigure}

    \caption{\textbf{The EAV24 race car}}
    \label{fig:example_racing}
\end{figure}

The vehicles used in the A2RL are EAV24s (Figure~\ref{fig:example_racing}), modified Dallara Super Formula SF23 cars tailored for autonomous racing.
The EAV24s have a modified \SI{2.0}{} liter turbocharged four-cylinder engine producing \SI{550}{} horsepower, a limited-slip differential, carbon brake disks, and Yokohama Advan racing slicks. An AMD Epyc 7313P Central Processing Unit~(CPU) with \SI{16}{} cores and an NVIDIA RTX 6000 Ada Graphics Processing Unit~(GPU) run the autonomous software. A Vectornav VN-310 Real-Time Kinematic Global Navigation Satellite System (RTK-GNSS) receiver with an integrated Inertial Measurement Unit (IMU) is available for localization. Furthermore, a Kistler SF-Motion provided an optical velocity measurement. For perceiving the other cars on the track, the EAV24 includes three Innovusion Falcon Kinetic Light Detection and Ranging~(LiDAR) sensors, four ZF Gen21 Radio Detection and Ranging~(RaDAR) sensors, and seven Leopard Imaging IMX728-9295-120H cameras.

\subsection{Related Work}

The task of autonomous racing represents a complex problem to solve for hardware and software. It has therefore attracted many researchers in the field of simulation racing \cite{wurman2022outracing}, drone racing \cite{kaufmann2023champion, hanover2024autonomous}, and car-like vehicles \cite{Betz2022}. Similarly to DARPA competitions \cite{tranzatto2022cerberus}, the community expanded with the creation of competitions for car-like vehicles—those provided events and platforms for evaluating autonomous driving software in a competitive racetrack environment. The competitions range from full-scale events, such as Roborace \cite{betz2019software}, the Indy Autonomous Challenge, and the Abu Dhabi Autonomous Racing League, to small-scale events, including Formula Student Driverless \cite{kabzan2020amz}, DeepRacer \cite{balaji2020deepracer}, and F1TENTH \cite{o2020f1tenth,baumann2025forzaeth}. %
Furthermore, high-performance sports cars or vehicle prototypes are used to demonstrate and validate research achievements for vehicle actions at the limit of performance and dynamics \cite{Funke2012, subosits2019racetrack, goh2024beyond}. Most of the research in autonomous racing focuses on advancements of individual autonomy algorithms, ranging from perception, planning, and control. 

\textbf{Perception:} Although the racetrack provides a less cluttered environment than an inner-city street scenario, the vast distance to the walls, the low landmarks given by monotonous surrounding objects, and the high speed make perception tasks quite challenging.  In the state of the art, most autonomous racecars are equipped with precise localization sensors, such as differential GPS (dGPS), that enable high localization quality. To cover GPS dropouts, tailor-made fusion techniques were demonstrated to allow for localization at high speed and in monotonous landmark environments \cite{Massa2020}. Although a few researchers have demonstrated visual localization techniques (e.g., SLAM) for autonomous racecars \cite{Gosala2019}, the field lacks dedicated developments for visual localization techniques that can achieve accurate localization at speeds above 200 km/h.
For object detection at high speeds above 200 km/h, a detection range of around 100m is indispensable, which can only be achieved through the multi-sensor fusion of camera, radar, and lidar. Previous research has not focused on dedicated object detection algorithm development and has instead used well-known state-of-the-art techniques like YOLO \cite{Strobel2020}. To further increase the development of new perception algorithms for racing, researchers created open-access datasets \cite{kulkarni2023racecar, vodisch2022}. 

\textbf{Planning:} With the knowledge of the vehicle’s changing dynamic state and the uncertain, interactive opponent behavior, another challenge is the motion planning at the limits of handling at and high speeds \cite{Jeong2013}. Ultimately, planning a trajectory in this adversarial environment involves calculating a fast and safe path between the opponent vehicles that maximizes progress on the track. As a solution to this problem, researchers provided graph-based \cite{Stahl2019_2, rowold2022efficient}, sampling-based \cite{Arslan2017,ogretmen2022smooth}, and optimization-based \cite{Reiter2021, Kapania2016} solutions for it. Consequently, a high re-planning frequency for the trajectory planning algorithm is indispensable for achieving real-time, low-latency autonomous maneuvers.

\textbf{Control:} Achieving human-like driving strategies when steering and accelerating the vehicle means implementing control strategies that can handle the car precisely at the vehicle's dynamic limits \cite{Kritayakirana2012_2}. It is crucial to ensure this accuracy across a wide dynamic range of tire temperatures, tire wear, weather conditions, variable friction, and bumpy track conditions. In the state of the art, it has already been demonstrated that model predictive control (MPC) is a highly successful method for computing control commands at high speeds with high accuracy \cite{Kloeser2020, Gandhi2021, Wischnewski2021}. Especially for racing, it is essential to guarantee robust control and, thus, the safe execution of the target trajectory in the presence of state noise and dynamic model uncertainties. Experienced race drivers, therefore, approach the limit of the car lap by lap while learning about the current dynamic setup of the vehicle and the condition of the race track. In an autonomous racecar, this can be achieved by applying learning-control techniques that can adjust the controller automatically over time \cite{Kabzan2019, Rosolia2020}. 

These research outputs demonstrate that the field of autonomous racing produces highly novel but isolated algorithms that have been tested either only in simulation or in lab environments. To address this gap, researchers have focused on creating holistic software architectures for real-world autonomous race cars~\cite{raji2024_1, raji2024_2, Saba2024-va, jung2023autonomous}. Here, the authors demonstrate how to design and develop modular and integrated software systems that operate robustly and competitively in real-world racing scenarios. %
The work presented in this paper builds upon the previous work of the TUM Autonomous Motorsport team~\cite{betz2023}. It presents the first fully integrated autonomous racing system that demonstrates consistent success in high-speed, multi-agent competitions by operating near the physical limits of the vehicle. Our main contributions are:

\begin{itemize}
    \item A modular software architecture for real-world autonomous racing;
    \item A robust perception system fusing LiDAR, radar, and GNSS for GNSS-degraded racing;
    \item A novel grip-aware spatial constraint representation used online and offline;
    \item A novel motion planning and control setup that handles multi-agent racing at 70 m/s;
    \item The team's deployment, testing, and race strategy;
    \item A real-world demonstration in the Abu Dhabi Autonomous Racing League (A2RL), winning the first four-car autonomous final ever held.
\end{itemize}

\section{TUM Approach}
\label{sec:tum_approach}

\subsection{Software Overview}
\label{sec:software_overview}

\begin{figure}[!btp]
    \centering
    \includegraphics[width=\textwidth]{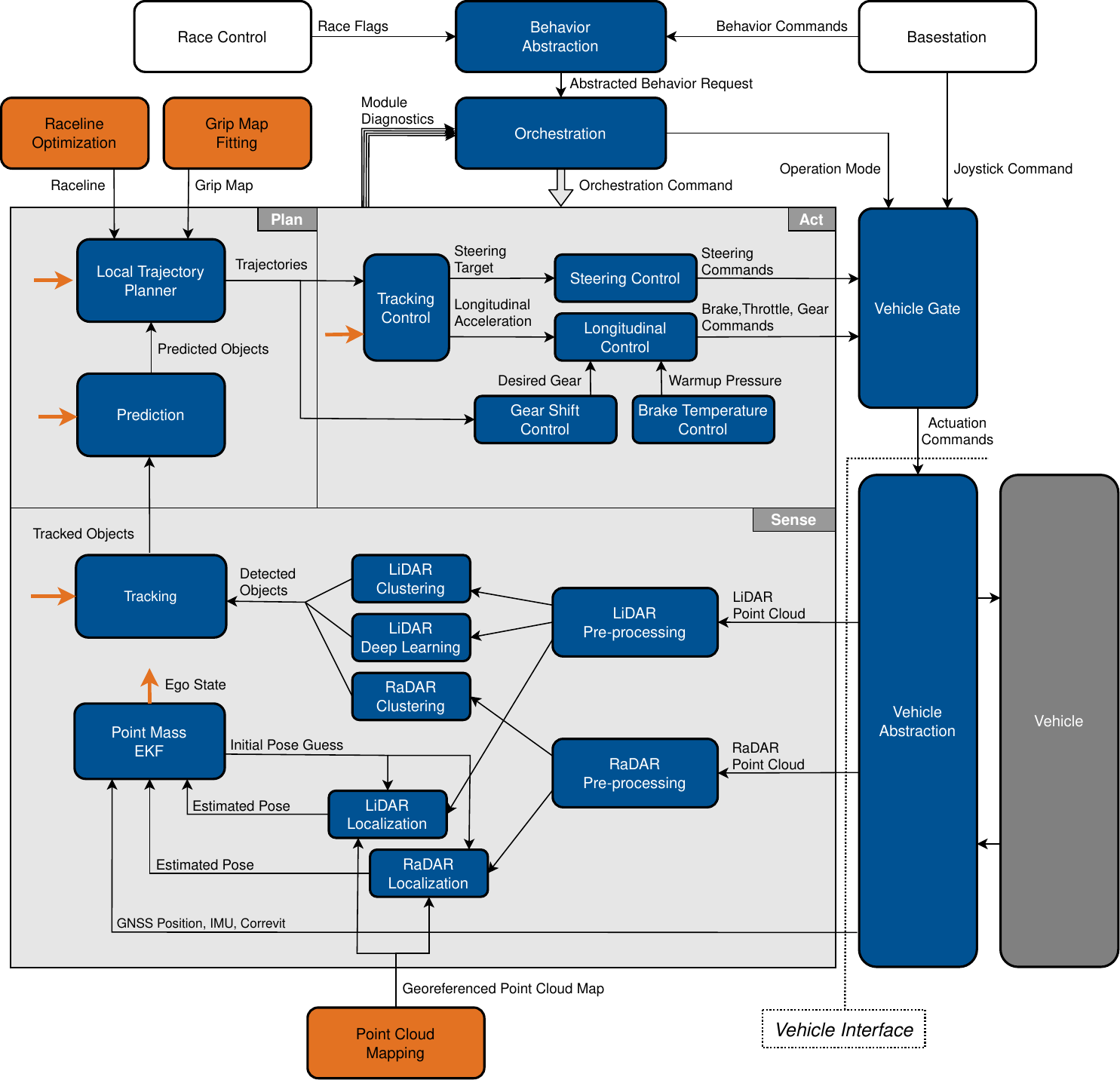}
    \caption{\textbf{Software architecture of TUM Autonomous Motorsport}. Blue components run on the vehicle during operation, orange components are executed offline, and white components run on a remote machine. For simplification, an orange arrow indicates modules that subscribe to the current ego state.}
    \label{fig:software-overview}
\end{figure}

Figure~\ref{fig:software-overview} provides a high-level overview of our software. At the top level, the stack consists of software running offline (orange), software running online on the car (blue), and software running online but not on the vehicle (white).

\textbf{Offline:} Raceline optimization focuses on generating an optimal raceline based on predefined assumptions. Another key offline component is the grip map, which allows for a spatial resolution of the acceleration constraints, enabling efficient testing and maximum performance in every corner. Furthermore, an offline-created three-dimensional point cloud map of the track enables precise localization even in GNSS-denied environments.

\textbf{Online:} According to the typical hierarchical robotics architecture, our core software splits into three subparts: \textsc{Sense}, \textsc{Plan}, and \textsc{Act}. \textsc{Sense} handles object detection and tracking and determines the ego vehicle's state. The map-based LiDAR-/RaDAR localization approach relies on a georeferenced point cloud map to accurately estimate the vehicle's location. This data, along with GNSS, IMU, wheel speed, and optical velocity (correvit) measurements, feeds into an Extended Kalman Filter~(EKF), which computes a three-dimensional ego state used by various modules. To accurately track other vehicles, we fuse the output of different object detection pipelines that operate on pre-processed LiDAR and RaDAR point clouds in a tracking algorithm.
The responsibility of the \textsc{Plan} portion of the software is to calculate both a target and an emergency trajectory using the ego state and all other tracked objects. Furthermore, the offline raceline, grip map, and orchestration commands (pit, overtaking allowed, speed limit, etc.) are required. Inside \textsc{Act}, the trajectory is executed by a hierarchical control structure consisting of a tracking controller and various low-level controllers. The vehicle gate forwards the controller's output to the vehicle and has the authority to overwrite commands in severe emergencies.
Race control and basestation can request certain behaviors from the vehicle, e.g., by race flags. These requests are processed by a state machine inside orchestration and forwarded to the core stack as orchestration commands. While the vehicle must always react to race control requests, basestation interventions are only allowed during testing. In addition to behavior requests, the state machine also monitors the health of each module to react to emergencies in individual modules.
Abstraction layers (vehicle and behavior abstraction) ensure that our core software remains independent of competition, vehicle, or simulation changes. These abstraction layers translate and transform inputs and outputs between competition or vendor specifics and our core software interface.
\\ %
The following sections provide a more detailed description of the most essential modules.

\subsubsection{Perception}
\label{sec:perception}

\begin{figure}[!btp]
    \centering
    \begin{subfigure}{0.49\columnwidth}
        \includegraphics[width=\textwidth]{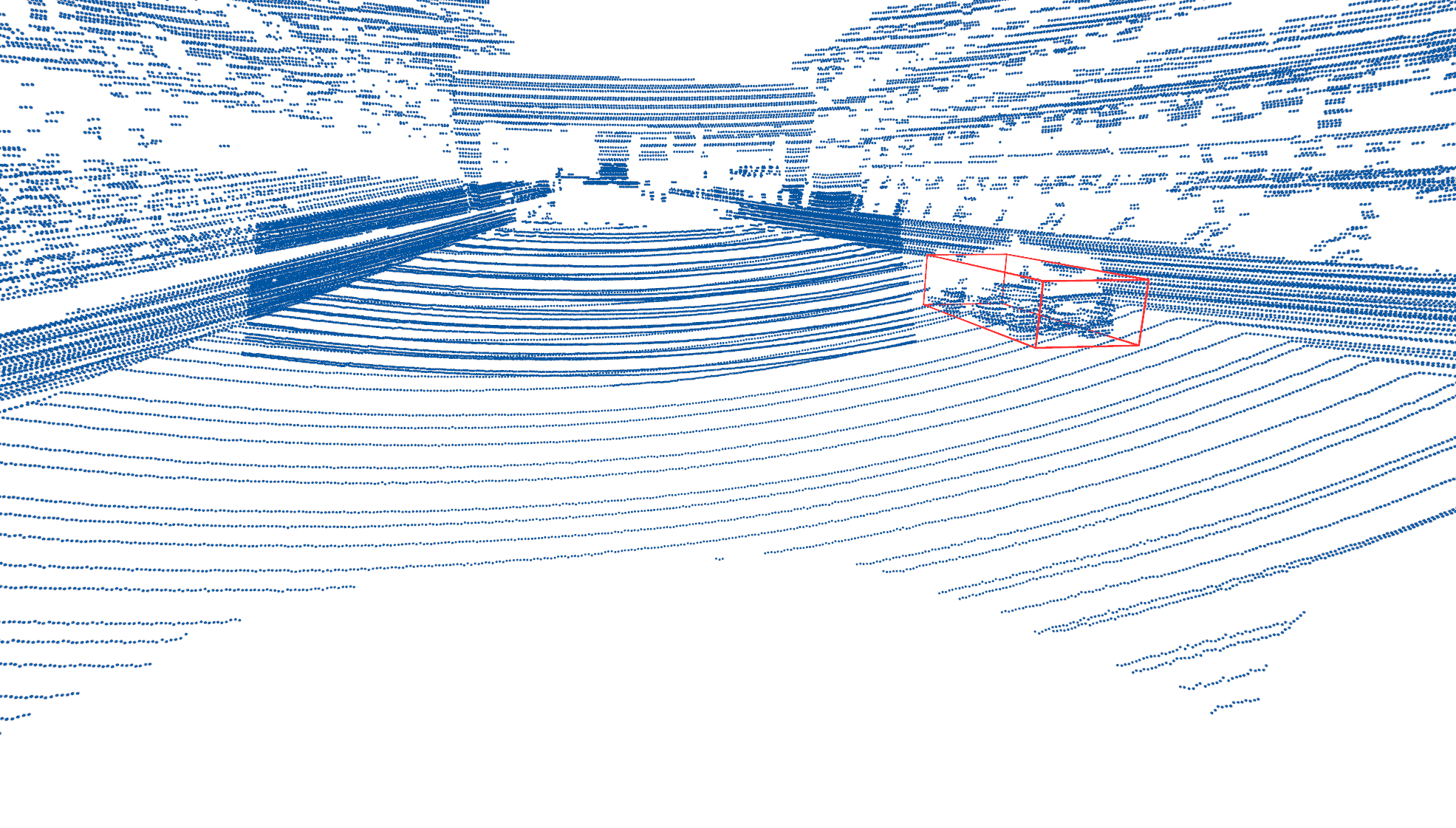}
        \caption{Full point cloud}
    \end{subfigure}
    \begin{subfigure}{0.49\columnwidth}
        \includegraphics[width=\textwidth]{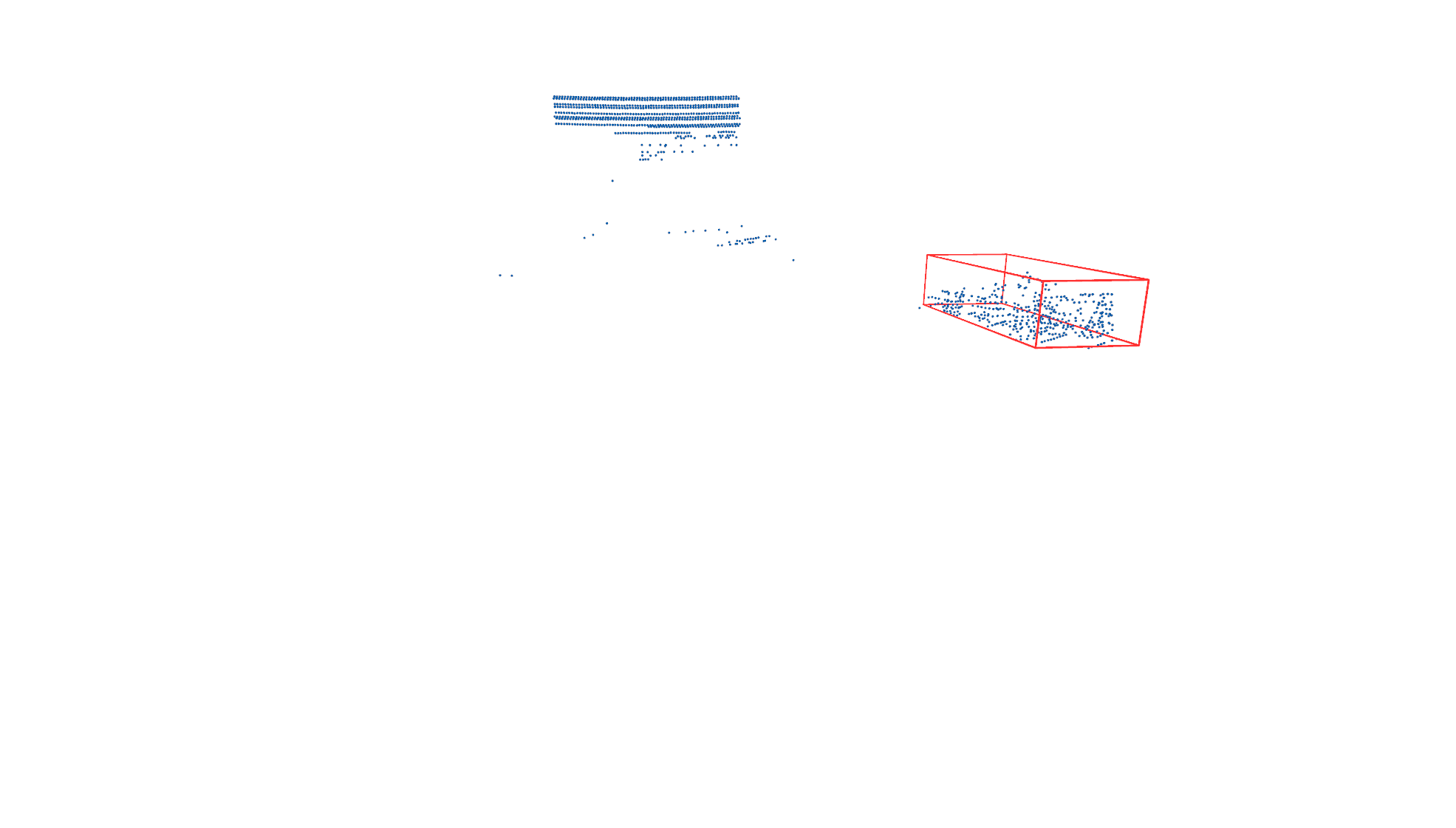}
        \caption{Filtered point cloud after pre-processing}
    \end{subfigure}

    \caption{\textbf{Comparison of the point cloud before and after pre-processing}. The ground truth annotation is visualized as a red box.}
    \label{fig:before_and_after}
\end{figure}

The perception pipeline is further divided into pre-processing, detection, and tracking.
To reduce processing times, a pre-processing stage reduces the size of the point cloud before any downstream detection and localization is conducted (Figure~\ref{fig:before_and_after}).

The point clouds from the individual LiDAR and RaDAR sensors, each covering dedicated areas around the car, are synchronized and concatenated into a single combined point cloud for each sensor type. Therefore, only one pipeline per sensor type is required, making the perception stack less dependent on different vehicle setups.
Our LiDAR pre-processing extends the approach presented by Betz et al.~\cite{betz2023} by adding additional filtering stages and improving the pipeline latency. This includes a track filter that removes points outside the drivable area and a threshold filter to filter dust, reflections, and sensor defects. 
Since our deep learning model has a fixed input size of \num{4000} points, the filter stages are tuned to reach this threshold (Figure~\ref{fig:before_and_after}). 
Compared to the original one containing \num{200.000} points, the filtered point cloud allows using smaller and faster detection models.
We also apply a subset of the pre-processing pipeline to RaDAR point clouds. For RaDAR pre-processing, the goal is to remove noise and outliers rather than reduce the size of the point cloud, as the RaDAR measurements are already sparse. For this reason, only points outside of track bounds and low-confidence measurements are removed.

LiDAR object detection is done redundantly by a clustering approach and a deep learning algorithm, as described by Betz. et al.~\cite{betz2023}. OpenPCDet~\cite{openpcdet} is chosen for the LiDAR deep learning framework as it allows quick comparison of many state-of-the-art algorithms with a feature-rich training pipeline.
We modified PointRCNN~\cite{shi2019pointrcnn} to have fewer parameters and use a smaller input point cloud.
This way, inference times shrink by a factor of 4 to \SI{19.8}{\milli\second} while yielding similar detection accuracy.
We use a RaDAR clustering algorithm to improve object tracking by supporting detection in areas not covered by LiDAR, such as blind spots caused by the rear wing.
Instead of clustering in three-dimensional Euclidean space like LiDAR, this algorithm detects objects based on x-y position and velocity measurements.
Detections from the LiDAR and RaDAR pipelines are fused and tracked over time using the tracking algorithm proposed by Karle et al.~\cite{karle2023multi}.
This presented perception stack consistently tracks dynamic objects at distances of up to \SI{200}{\meter} and velocities of up to \SI{70}{\meter\per\second}.

\subsubsection{Localization and Vehicle State Estimation}
\label{sec:localization}
\begin{figure}[!btp]
    \centering
    \includegraphics[width=\textwidth]{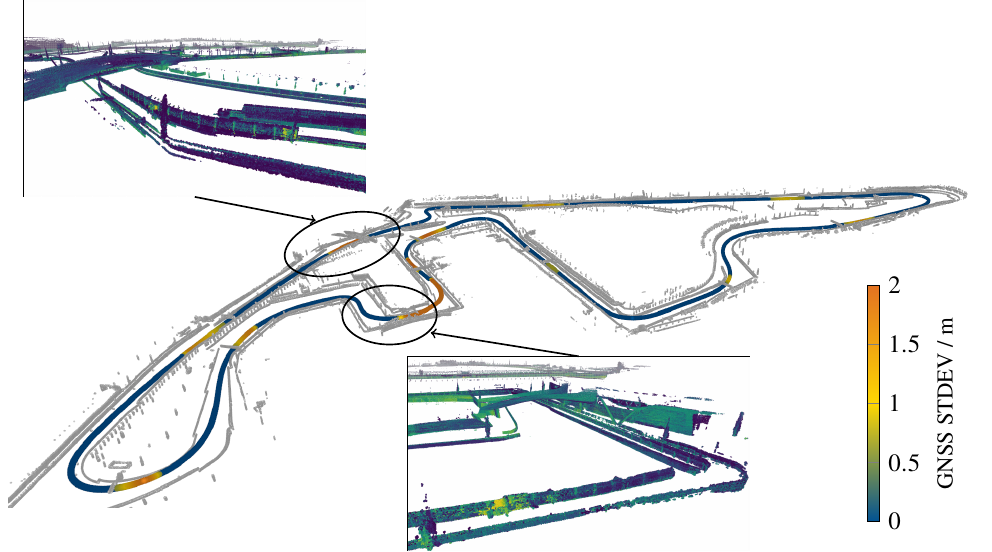}
    \caption{\textbf{Illustration of the used static point cloud map}. The coloring of the drawn trajectory indicates the GNSS standard deviation (STDEV) on the different parts of the race track.}
    \label{fig:localization_pointclouds}
\end{figure}

Robust estimation of the vehicle's pose and its dynamic state, consisting of the vehicle velocities and accelerations, forms a core component of an AD software stack (Figure~\ref{fig:software-overview}). However, the demanding racing conditions present several challenges, including intense vibrations caused by the combustion engine and the stiff chassis. This requires high IMU sampling frequencies and customized filter coefficients to prevent aliasing effects. Furthermore, missing GNSS coverage on multiple sections of the track~(Figure~\ref{fig:localization_pointclouds}) requires multi-modal sensor fusion, leveraging the individual strengths of each sensor. \\
We employ a three-dimensional EKF to estimate the vehicle's pose and dynamic state~\cite{svenster}. A map-based localization approach utilizing LiDAR and RaDAR  data accounts for the limited GNSS coverage around the track. Point cloud mapping is conducted offline and involves a two-stage pipeline: First, we generate an initial map by combining an odometry trajectory from \textit{KISS-ICP}~\cite{kiss_icp} with manually inserted loop-closure constraints using \textit{Interactive SLAM}~\cite{koide}. In the second stage, the constructed map is georeferenced using the method outlined by Leitenstern et al.~\cite{flexcloud} to ensure alignment with GNSS coordinates.
The online LiDAR and RaDAR localizations calculate a pose in the local reference frame by aligning the measured LiDAR and RaDAR point clouds with the pre-built map. The iterative point cloud registration pipeline employs a modified version of the \textit{KISS-ICP}~\cite{kiss_icp} algorithm, using the state estimate of the EKF-based sensor fusion as the initial guess. The calculated pose is then fed back into the EKF. To mitigate delays introduced by the pre-processing and registration stages of the pipeline, we compensate for these delays using the Kalman Filter's state history. Consequently, precise synchronization between the sensors and low execution times are crucial for robust localization, making the low-latency RaDAR localization a key component.

The map-based localization approaches achieved an accuracy comparable to the RTK-corrected GNSS positions, making them a reliable alternative in the event of GNSS dropouts. The measurement noise covariance of each localization source (GNSS, LiDAR, RaDAR) is dynamically adapted based on the reported GNSS and estimated point cloud registration uncertainty.

The overall concept delivers robust estimation of the vehicle's pose and dynamic state, preventing sudden jumps during sensor failures or GNSS dropouts and ensuring smooth control under harsh conditions.

\FloatBarrier %
\subsubsection{GripMap: Spatially Resolved Vehicle Dynamic Constraints}
\label{sec:grip_map}

In real-world racing, both surface properties and the overall ability of the driver or autonomous software to fully exploit the vehicle's dynamic potential can vary substantially across the track. For instance, debris or rubber buildup leads to localized grip changes that a single, global vehicle model cannot represent. As demonstrated by Werner et al.~\cite{werner2025gripmapefficientspatiallyresolved}, spatially resolved models allow for better performance where the grip is higher while remaining safe where friction is limited.

\textbf{Motivation.}
A key motivation for using spatially resolved constraints is that tire friction potential varies depending on the location around the track. Measurements of multiple racetracks have shown that grip potential can fluctuate significantly due to changing surface conditions, with variations of \SI{50}{\percent} being common and, in some cases, even more~\cite{Woodward.2012, Wadell.2019}.

\textbf{Method.}
We discretize the track into a grid corresponding to the longitudinal $s$ and lateral $n$ raceline-relative positions to account for this. Each of these Frenet-cells stores a dimensionless multiplier \(\theta_{ij}\). As illustrated in Figure~\ref{fig:offline_raceline}, these multipliers adjust the baseline vehicle-dynamics constraints locally, reflecting differences in grip. In practice, \(\theta_{ij}\) can be obtained or refined through track data, telemetry, and iterative post-run analyses. The \(\theta_{ij}\) lookup operates in constant time by hashing \((s,n)\) into a 2D array, making it computationally efficient and, therefore, suitable for real-time high-frequency planning applications.

\textbf{Baseline Constraints.}
We use a point-mass model for the baseline vehicle constraints, which is advantageous due to its low computational demand. The model is constrained by maximum longitudinal and lateral accelerations, extended to account for aerodynamic effects and 3D track geometry by including velocity \(v\) and vertical acceleration \(a_z\) as independent variables. We derive these baseline g-g-g-v constraints via the quasi-steady-state black box simulation approach presented by Werner et al.~\cite{werner2025quasisteadystateblackboxsimulation}, using a validated two-track model and thus avoiding model mismatch. The local multipliers \(\theta_{ij}\) then refine these global limits:
\[
    a_{x,\max}(s,n) \;=\; \theta_{ij}\,\tilde{a}_{x,\max}(v,a_z),
    \quad
    a_{y,\max}(s,n) \;=\; \theta_{ij}\,\tilde{a}_{y,\max}(v,a_z),
\]
where \(\tilde{a}_{\square,\,\max}\) denotes the globally derived g-g-g-v values.

\textbf{Offline and Online Planning.}
When calculating the time-optimal raceline offline, the local scaling factors \(\theta_{ij}\) allow the optimization to exploit sections with higher friction potential while ensuring safe limits in areas with lower grip. \\
In online planners, each possible trajectory is checked against these local constraints. If a trajectory requires more acceleration than \(\theta_{ij}\) permits, it is either penalized or discarded to maintain feasibility. Additionally, GripMap allows assigning lower grip values away from the main raceline. This reflects increased uncertainty in areas of limited data availability and possible debris buildup, leading to decreased grip potential in these zones. As a result, the vehicle behaves more cautiously in multi-vehicle maneuvers or overtakes that deviate significantly from the raceline.

\subsubsection{Offline Raceline Optimization}
\label{sec:racelineplanner}

\begin{figure}[!btp]
    \centering
    \includegraphics[width=1.0\columnwidth]{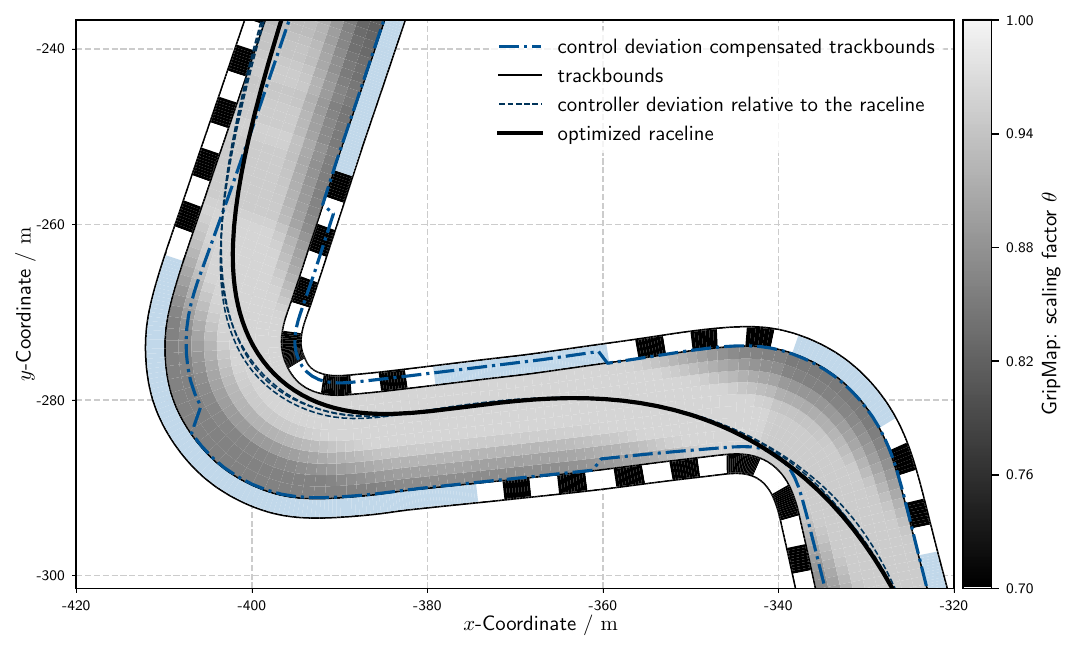}
    \caption{\textbf{Time-optimal raceline through turns 6 and 7}. To improve visibility, the controller induced lateral offset to the racing line is increased by a factor of 2 in this figure compared to the real world. The grayscale grid shows the spatially resolved acceleration limits used for optimization and online planning.}
    \label{fig:offline_raceline}
\end{figure}

Many tasks in autonomous driving must be executed in real time. However, precomputing certain information offline reduces the computational burden. In addition to the point cloud map and grip map mentioned earlier, we calculate the optimal raceline and the racetrack representation offline. Multiple modules, such as local planning, grip map generation, and prediction, operate in curvilinear coordinates, requiring a three-dimensional reference frame. This reference frame~\cite{limebeer2015} is generated around the offline-optimized, time-optimal raceline.

Finding a balance between smoothness and accuracy prevents effects like oscillatory vehicle behavior caused by the track representation. 
To achieve this, we developed a novel pre-processing method, which takes independent raw cartesian points of the track bounds, filters these with tangential polynomial resampling, transforms each to spatial-frequency space, low pass filters them, and finally samples the intersections of these continuous back-transformed functions with the orthogonal plane of the center line at the desired discretization intervals.

The optimization method, as well as the representation of the vehicle limits, is based on Rowold et al.~\cite{rowold2023} and Lovato and Massaro~\cite{lovato2022} with the extension and modifications described in Section~\ref{sec:grip_map}.
The optimization is an Optimal Control Problem (OCP), which is converted into a nonlinear program through explicit Runge-Kutta collocation. The vehicle's dynamic limits are handled according to Section~\ref{sec:grip_map}, but as these limits only represent the quasi-steady-state behavior, they do not contain transient effects.
Furthermore, a s-based maximum velocity constraint allows for more flexibility and fine-tuning of the racelines.

Due to modeling errors and algorithmic limitations, the vehicle will inevitably deviate from the target trajectory sent to the control module (Section~\ref{sec:control}) when approaching the dynamic limit. A portion of these deviations is repeatable lap-to-lap, which enables the compensation of these deviations in offline optimization by shifting the track bounds locally using the s-based solver constraints. Figure~\ref{fig:offline_raceline} shows the control deviation in turns 6 and 7 and the track bound compensation through local constraints.

Furthermore, the weight transfer caused by a short longitudinal acceleration between successive corners introduced transient effects, reducing control performance. Velocity caps are introduced in certain parts of the track to account for those effects.

\subsubsection{Prediction}
\label{sec:prediction}

The tracked objects, provided by the \textsc{Sense} stack, are fed into the prediction module to anticipate the further movement of surrounding objects. We send one predicted trajectory per opponent to the planning algorithm. The motion prediction, in part, relies on a prediction raceline calculated according to Section~\ref{sec:racelineplanner}. 
This assumption is reasonable, as all competitors aim to achieve the fastest lap around the track.

\begin{figure}[!bt]
    \centering
    \includegraphics[width=1.0\columnwidth]{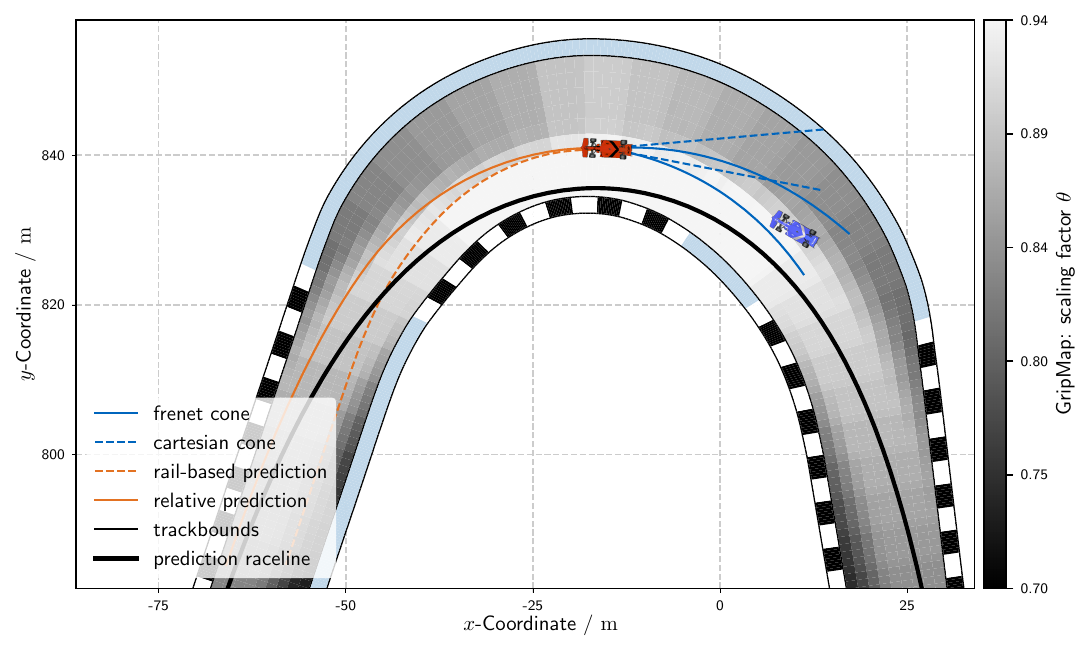}
    \caption{\textbf{Exemplary prediction of a leading vehicle in turn 5 of the Yas Marina Circuit}. The solid line shows the relative, and the dashed line the rail-based prediction method. The cartesian cone is indicated with a dashed line, while the Frenet cone is drawn solid. The prediction of the orange vehicle is visualized from the perspective of the blue car. From the perspective of the orange car, we would not react to the blue vehicle since it is inside the virtual cone.}
    \label{fig:prediction}
\end{figure}

The opponent's relative position between the prediction raceline and the track boundary (Figure~\ref{fig:prediction}) supports predicting its lateral position. This relative position is a scaling factor for the whole prediction horizon. If the opponent is initially located at \SI{40}{\percent}, while \SI{0}{\percent} means on the raceline and \SI{100}{\percent} on the corresponding track bound, the lateral position stays at \SI{40}{\percent} for the whole prediction horizon.

Similar to the lateral movement, the relative velocity between the actual opponent's state and the velocity of the prediction raceline yields a scaling factor for the whole prediction horizon.

While the proposed concept can handle minor deviations from the prediction raceline, a significant deviation indicates a driving behavior not covered by the prediction raceline. 
Those cases, e.g., during an overtaking maneuver, are handled differently by considering the ruleset of the event:
\begin{itemize}
    \item The leading vehicle has the right-of-way and can follow its raceline until the trailing car initiates an overtaking attempt. An overtake starts if the trailing car crosses defined lateral and longitudinal offset thresholds to the leading car. This does not mean that the leading car has to open the gap, but it has to leave enough space if it was available when the attempt started. In those cases, we assumed an opponent vehicle in front would hold the line. Hence, it maintains its absolute distance from the track bounds.
    \item To detect when to grant right-of-way to a trailing opponent, a virtual cone behind the ego vehicle is specified in curvilinear coordinates (Figure~\ref{fig:prediction}). Using this cone, rear opponents are only considered if they are close and have a sufficient lateral offset (overtaking attempt). Cone dimensions correlate with the engaging distances specified by the right-of-way ruleset.
\end{itemize}

\subsubsection{Local Trajectory Planning}
\label{sec:planning}

To enable safe and strategic multi-vehicle interactions, such as overtaking opponents and avoiding collisions, we employ a sampling-based local trajectory planning approach inspired by Ögretmen et al.~\cite{ogretmen2024}, designed to account for three-dimensional track effects. Trajectories are generated with a fixed-time horizon of \SI{4}{s}, balancing long-term planning and maneuverability. To connect the start state with the sampled end states, we utilize fourth-order polynomials for the longitudinal and fifth-order polynomials for the lateral trajectory. The longitudinal sampling strategy integrates a hybrid approach, generating purely jerk-optimal trajectories based on Werling et al.~\cite{werling2010} and raceline-relative trajectories. The jerk-optimal solutions are used to achieve smooth speed transitions and steady-state driving. Relative trajectories enhance adaptability to road courses by embedding features of the offline-optimized time-optimal raceline described in Section~\ref{sec:racelineplanner}, such as brake points, directly into the velocity profile. Additionally, we increase sampling density near the raceline to accommodate minor deviations necessary for operating at the vehicle dynamics limit.

Trajectory feasibility is assessed through a four-stage validation process. First, curvature constraints ensure compliance with the vehicle's minimum turning radius. Second, trajectories that extend beyond track boundaries are discarded. Third, velocity limits imposed by race control are strictly enforced. Fourth, adherence to the acceleration constraints is verified, with a small margin permitted to prevent premature rejection of near-feasible trajectories. These constraints are further adapted using the GripMap (Section~\ref{sec:grip_map}), allowing the planner to account for local variations in surface grip.

\begin{figure}[!btp]
    \centering
    \includegraphics[width=1.0\columnwidth]{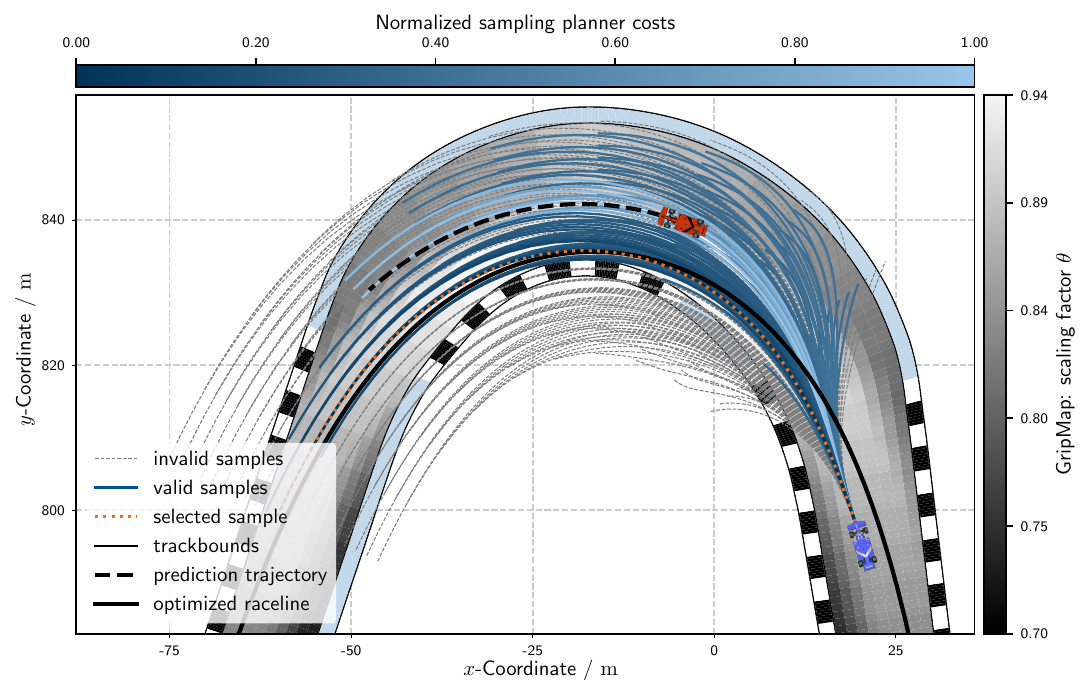} %
    \caption{\textbf{A set of trajectory samples generated by the local planning algorithm.} Samples grayed out are invalid, and the cost of the valid samples is indicated by color for a mock cost function. This scene is a recreation of the winning overtake into turn 5 during the grand final.}
    \label{fig:samples}
\end{figure}

Trajectory selection relies on a cost function composed of six weighted terms, each tuned to optimize competitive driving performance. The function considers lateral deviation from the raceline, velocity tracking error, deviation from raceline curvature, minor acceleration limit violations, collision risk, and proximity to other vehicles' predictions. The latter three terms play a crucial role in multi-vehicle interactions. For example, while acceleration limit violations are generally undesirable, they may be acceptable if they help to prevent a collision with another vehicle. To enhance adaptability in multi-vehicle racing, we model a dynamic elliptical safety region around competing vehicles, which modifies the proximity costs. Furthermore, we scale the size of this ellipse with increasing speed to accommodate varying risk levels.
Figure~\ref{fig:samples} shows a recreation of the winning overtake during the A2RL grand final, displaying the sampled trajectories, their feasibility, and their costs.

\FloatBarrier
\subsubsection{Control}
\label{sec:control}

The motion control algorithm is responsible for determining the optimal actuation signals to safely follow the trajectory provided by the local planner. The utilized approach extends our proven hierarchical control architecture~\cite{betz2023}, consisting of a high-level Tube Model Predictive Controller (TMPC)~\cite{Wischnewski2022TubeMPCApproachHighSpeedOvals} and low-level controllers for longitudinal~\cite{pitschi2025} and lateral dynamics.
The high-level TMPC calculates a feasible trajectory of lateral and longitudinal accelerations considering the vehicle state, model, and constraints.
The low-level lateral and longitudinal controllers realize the requested accelerations and account for the dynamics not modeled in the TMPC via a cascaded PID control structure while heavily relying on feedforward control.

\begin{figure}[!btp]
    \centering
    \includegraphics{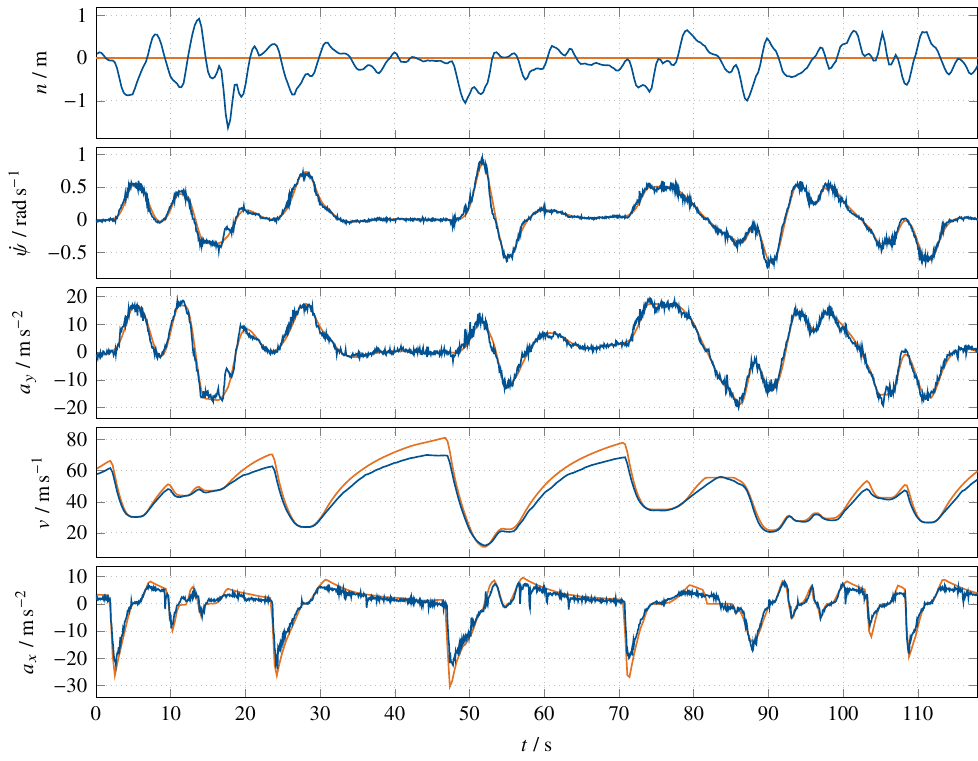}
    \caption{\textbf{Vehicle dynamics control behavior on our fastest lap during prequalification}. The planning target is depicted in orange, while the blue color corresponds with the measure value. }
    \label{fig:control_behavior}
\end{figure}

Figure~\ref{fig:control_behavior} shows the control behavior during our fastest lap. Except for turn 3 ($t=18$), our approach stayed below a lateral deviation $n$ of 1 meter to the planned trajectory while reaching lateral $a_y$ and longitudinal accelerations $a_x$ of around \SI{20}{\meter\per\second\squared}. We reached a maximum velocity $v$ of approximately \SI{70}{\meter\per\second} and yaw rates $\dot{\psi}$ of up to \SI{0.9}{\radian\per\second}.

In previous works~\cite{Wischnewski2022TubeMPCApproachHighSpeedOvals}, our TMPC algorithm used a diamond shape to describe the coupled acceleration limits.
However, this approach heavily underapproximates the dynamic potential of the car. Switching to an octagonal representation allows for following trajectories that are planned more closely to the actual limits of the vehicle, especially in combined acceleration situations. This way, we could significantly improve the lap time we achieved.

Since our localization system fuses signals from multiple sensors, small drifts can occur, particularly when the validity of one of the inputs changes.
To mitigate the impact of such drifts, we reparameterized the control algorithm to maintain lower lateral deviations from the planned trajectory than the parameter set originally optimized for high-speed oval driving. This increases the safety margin to handle localization-induced virtual deviations while satisfying the hard constraints defined by the driving tube, allowing for a smoother control reaction. As a result, we achieved both improved tracking performance and enhanced numerical stability in the optimal control problem.

To safely operate near the vehicle's dynamic limits, smooth and robust actuation of longitudinal velocity is highly important. Therefore, we integrated traction control (TC) and anti-lock braking (ABS) systems into the low-level control architecture to correctly react in case we oversaturate the dynamic potential of the tire. These systems use a fixed slip threshold to remain independent of tire models and enable rapid intervention for high tire slip values via brake pressure adjustments~\cite{pitschi2025}.

Additionally, we implemented a predictive gear shift controller to prevent drivetrain-induced disturbances and the resulting instabilities during cornering. It forecasts engine speeds along the planned trajectory and proactively shifts before corners.
Furthermore, we implemented a brake warmup controller that accelerates the heating of the highly temperature-dependent carbon-ceramic brakes by constantly applying a light brake pressure. It shortens the warmup phase of the car and, therefore, results in more effective testing time~\cite{pitschi2025}.

\subsubsection{Orchestration}
\label{sec:orchestration}

To safely operate the core functionalities from \textsc{Sense} to \textsc{Act}, the orchestration supervises all software components via a state machine and a watchdog. Each module reports its status and reacts to internal or external faults, while the watchdog ensures that the diagnostic state is regularly updated.  The module states are communicated using \ros{} diagnostic messages containing key metrics and one of four diagnostic levels (\textsc{Ok}, \textsc{Warn}, \textsc{Error}, \textsc{Stale}).

States and key metrics are visualized on a dashboard for human oversight. However, based on the module and the associated error level, the system automatically triggers one of the following emergency responses~\cite{sagmeister2025}:
\begin{itemize}
    \item \textbf{Safe Stop.} The vehicle decelerates to a stop while still actively replanning its trajectory.    
    \item \textbf{Emergency Stop.} The vehicle switches to the last safe emergency trajectory, decelerating to a stop as fast as possible.
    \item \textbf{Hard Emergency.} Driving autonomously is no longer safe, so the car applies all brakes and sets the steering straight.
\end{itemize}

During nominal operation, external requests can be sent by race control or the basestation. 
Each flag received is translated into a behavior request.  Combining basestation and race control requests, the state machine generates and distributes an orchestration command within the core stack. This orchestration command includes, for example, maximum velocity, overtaking permission, following distance, and a request to pit.

\subsection{Deployment}
\label{sec:deployment}

The short lead times to the final race require efficient deployment strategies for the developed algorithms. As presented in previous publications~\cite{betz2023}, we divide our software into containerized microservices. The microservices provide a specific functionality each, which can solve the AD task when combined.
The microservice-based approach greatly improves testing efficiency since the same binaries run in simulation and during real-world operation. A Continuous Integration/Continuous Deployment (CI/CD) pipeline automatically builds the microservices from any code update.

We reduced the size of the Docker images for each microservice by using Docker's layer caching. Copying locally built binaries into an otherwise prebuilt image ensures that only the last layer of the image is updated. As a result, downloading an updated container image is just a download of the new module binaries. The small download size simplifies updates and increases testing efficiency.

In addition to using containers for abstracting the computing hardware, we abstract the whole vehicle by using a generic vehicle interface as a portal into our software.
This interface needs to be supported by all vehicles and all simulators. Therefore, individual abstraction layers translate from the vendor to the vehicle interface. These abstractions simplify the deployment to a new vehicle since the hardware drivers and their abstraction only need to provide this predefined list of topics~\cite{sagmeister2025}.

Despite limited prior information, these concepts allowed us to drive a new car autonomously within a few days.

\subsection{Testing Strategy}
\label{sec:testing_strategy}

\subsubsection{Simulation}
\label{sec:simulation}

Before real-world testing, we extensively simulated the entire software stack to assess new features and overall performance. We developed a Software-in-the-Loop~(SIL) setup for multi-vehicle racing, where multiple machines run instances of our stack in parallel.
Our basestation manages the individual software stacks, replicating the real-world operation of the vehicle on the race track.

The base simulation software shows a modular architecture. At its core, a high-fidelity vehicle dynamics model~\cite{sagmeister2024} provides an accurate simulation of vehicle behavior. We enhanced this vehicle dynamics model by incorporating a limited-slip differential as well as brake and tire temperature dynamics. 
This allowed us to utilize simulation to test our temperature management strategies, which played a crucial role in the competition.

To test the multi-vehicle capabilities of the software stack, we conduct simulations using multiple instances of our software. 
The simulation provides the vehicle states of each agent to all other agents as virtual detection.
This approach requires minimal computational resources and allows for flexible connectivity between various machines, even basic laptops, to test interaction scenarios. This was vital given the race track's lack of simulation infrastructure and limited network bandwidth.

Furthermore, we developed a complete high-fidelity sensor simulation. This is essential for simulating perception and localization, allowing us to test closed-loop multi-vehicle behavior with the full software stack instead of exchanging ground-truth positions. The sensor simulation includes LiDAR, RaDAR, Camera, and CAN bus simulations, all modeled to match the actual vehicle.
The full-stack simulation also helps to evaluate deployment strategies on the HIL test bench, which features a computing platform identical to the vehicle. By simulating real-world workloads, we could test resource allocation strategies (e.g., CPU pinning) and assess their impact on software performance.

Before each test day, we conduct a complete dry run of all planned tests in simulation. This allows for anticipating the vehicle's expected behavior during real-world testing and quickly identifying deviations. Additionally, it serves as a final quality gate for new features, preventing crashes and unnecessary loss of valuable track time.

\subsubsection{Real-World Testing}
\label{sec:real_world_testing}

The previously described simulation environment provides a key tool for fast development, baseline validation, and parametrization. However, real-world environments entail additional challenges that are often not sufficiently covered, even in high-fidelity simulators.

First, we evaluated the ABS and TC systems, which should later provide a safety net when increasing performance. Those were tested on straight lines or in turns with large runoff zones while continuously increasing lateral and longitudinal accelerations.

With the safety net tested, we aimed to optimize single-vehicle performance. We incrementally increased our target acceleration to gather insights into the vehicle limits, first in low-speed corners and in turns with large runoff zones.
This risk-aware testing strategy prevented excessive costs and improved the overall vehicle condition. Each accident poses additional stress to all components, increasing the risk of further failures in the future.

One key factor in the competition was starting on cold tires, which warmed up during each stint. Race car tires mark a narrow temperature window, yielding degraded performance at lower temperatures. Therefore, we considered the previously tested lateral accelerations at different tire temperatures and velocities for each test to make a conservative assumption about the vehicle limit.

In addition to temperature sensitivity, the performance of our control algorithm was the limiting factor in various turns.
We used the introduced GripMap to limit and gradually increase lateral acceleration in particularly challenging corners, and we step-by-step fine-tuned our controller to improve overall performance.

Besides the timed runs, multi-vehicle interaction was central to the competition. Therefore, perception and prediction had to be evaluated using real-world data. We started running with other teams on split lines with overtakes at gradually increasing speeds to gather a dataset for our perception algorithms. Next, we ran same-line overtakes, again with increasing speeds.

\section{Results - The Race}

As previously mentioned, the competition consisted of three individual events.
These events included challenges beyond those encountered in previous full-scale autonomous racing competitions~\cite{betz2023}. The road course enabled higher longitudinal accelerations combined with sharp, alternating turns, increasing the importance of transient vehicle dynamics compared to oval racing.
Additionally, the potential for significant delta speeds between vehicles requires low-latency and long-range opponent detection to ensure timely and effective decision-making. Other challenges included extreme temperature variations and the first-ever four-vehicle autonomous race, further pushing the limits of perception, planning, and control.

Consistently exploring the dynamic limits at different ambient and tire temperatures was crucial for our final racing strategy. Based on this testing data, we developed a tire temperature model, allowing us to integrate tire warm-up behavior into trajectory planning, maximizing performance throughout the expected temperature range. Furthermore, utilizing the introduced grip map enabled spatially resolved limits to cope with available tire grip and controller behavior. 

This strategy led to a final lap time of \SI{118,985}{\second} during the pre-qualification, which determined the seeding for the attack-and-defend competition. This final lap time was the second fastest among all competitors.
The performance during the two-month testing phase and the final event showed remarkable improvements, leading to a final gap of less than \SI{10}{\percent} compared to the professional racing driver Daniil Kvyat.

During the attack-and-defend competition, an overtake attempt resulted in a crash with the opponent's vehicle. Our software anticipated that the opponent would aim for the apex and planned to follow behind them through turns 6 and 7. However, they started braking earlier than expected, finally leading to a crash between the cars. This still meant a second place in the attack-and-defend competition since we were seeded directly in the final.

Due to technical issues during the time-trial, we had to start from the third position during the final four-vehicle race. This posed additional pressure as two opponents had to be overtaken to win the competition.
Consequently, we conducted extensive simulations on various interaction parameters in different overtaking scenarios to avoid an incident similar to the attack-and-defend competition. 
During the race, the grip map, combined with the tire-temperature model, ultimately allowed for faster driving at lower tire temperatures than the opponent vehicles. This enabled our car to quickly close the gap to the better-qualified teams. Subsequently, our software stack successfully attempted the final overtake in turn 5 on the inner line (Figure~\ref{fig:samples}), ultimately securing the victory.

\section{Discussion and Conclusion}

In this work, we presented our approach to successfully deploy and tune our software on a full-scale autonomous racing car within two months, achieving performance within \SI{10}{\percent} of a professional human driver. This effort ultimately led to our victory in the inaugural A2RL. We attribute this success to five key enablers:
\begin{description}
    \item[Grip Map] The grip map allowed us to push harder in certain corners while we were already at the limit in others. Additionally, it enabled us to integrate track-specific knowledge gained from test runs, maximizing our software's performance across the entire track.

    \item[Stability Control] To refine the grip map, we progressively pushed the vehicle to its dynamic limits, corner by corner. Taking too small steps wastes valuable track time, and too big steps risk exceeding the limit and losing the car. The low-level stability control systems (TC, ABS) allowed us to test quickly and safely.

    \item[Redundancies] Even though the software performs a safe stop in the event of hardware or software failure, such incidents still result in lost track time. Implementing backup layers (LiDAR and RaDAR localization) enabled us to complete test runs despite certain shortcomings.

    \item[Understanding Tire Temperature Behavior] The race format did not allow for dedicated tire warm-up, meaning a considerable portion of the event was run on cold tires. As a result, we needed to collect enough data during testing to enable fine-tuning our software to the tire behavior over a broad range of tire temperatures, ensuring both safety and competitiveness during the event.

    \item[Architecture] Given the limited time available with the car and the evolving nature of the competition, we designed our software architecture to be highly flexible. This adaptability allowed us to conduct a successful test run on the very first day, rapidly integrate new features we identified as crucial during testing (such as the brake temperature controller, gear shift, ABS, and TC), and quickly adjust to changes in race format (e.g., flag response behavior).
\end{description}
Finally, we would like to share our key findings on the current limitations of the state-of-the-art AD algorithms, based on our experience in this competition.

During the attack-and-defend run, the crash highlighted a critical limitation: relying on a single predicted trajectory is insufficient to capture the full variability of opponent behavior. Uncertainties in opponent decisions - such as whether they yield - must be accounted for. This, by concept, cannot be done with a single prediction trajectory since every possible trajectory inside the planning algorithm would influence the opponent's plan in the future.
However, game-theoretic approaches could capture these interdependencies in the future.

Another key insight concerns our simulation methodology. Further automation must account for the vast variability of racing scenarios for highly interactive maneuvers. Moreover, our multi-vehicle simulations posed the risk of overfitting, as all tested scenarios involved agents running the same software stack. Even with parameter modifications to introduce behavioral variability, specific inherent characteristics of our software remained deeply embedded. Interacting with opponents using different base software stacks could expose unforeseen edge cases. To address this, a shared simulation environment across teams could help to generate richer interaction data.

The manual effort required to generate the grip map was significant. To reduce this burden, the vehicle should dynamically create the grip map in real-time based on various performance metrics, allowing it to learn and adapt to the limit.

While quasi-steady-state vehicle models were sufficient to achieve performance within \SI{10}{\percent} of a human driver, transient vehicle dynamics become increasingly important when pushing performance further.

Finally, planning at the limits of dynamic feasibility, coupled with the complexity of road course geometry and uncertainty in opponent predictions, introduced challenges in ensuring the recursive feasibility of planning. Addressing these issues will be crucial for advancing autonomous racing performance.

Nevertheless, expanding the safety envelope with autonomous racing by demonstrating increased safety through agility can generate further trust for AD. Similar to what we have seen in the past with IBM's DeepBlue or DeepMind's AlphaZero, we can strengthen our confidence in autonomous vehicles when we one day demonstrate an autonomous race car competing and winning against the current Formula 1 champion.

\clearpage %

\section*{Acknowledgments}
We gratefully acknowledge all members of TUM Autonomous Motorsport for their dedication and contributions. Special thanks go to the current team for their ongoing efforts, to past members for sharing their experience and providing a solid foundation of code, and to the student members whose thesis work has significantly advanced our software stack and introduced innovative concepts.
We also extend our sincere appreciation to all sponsors and partners of the team for their continued support. Finally, we thank the organizers of the Indy Autonomous Challenge (IAC) and the Abu Dhabi Autonomous Racing League (A2RL) for providing platforms that enable real-world testing and continuous development of our autonomous racing software.

\paragraph*{Funding:}
\begin{itemize}
    \item J. Bongard and S. Sagmeister were funded by the Deutsche Forschungsgemeinschaft (DFG, German Research Foundation) – 469341384
    \item C. Schröder was funded by the Bayerische Forschungsstiftung (Project Ref. AZ-1591-23)
    
\end{itemize}
\paragraph*{Author contributions:}
S. Hoffmann and S. Sagmeister, as the first authors, designed the structure of the article and contributed essentially to 
the overall concept and architecture of the software, as well as the implementation of various modules and libraries.
J. Betz
T. Betz,
J. Bongard,
S. Büttner,
D. Ebner,
D. Esser,
G. Jank,
S. Goblirsch,
A. Langmann,
M. Leitenstern,
L. Ögretmen,
P. Pitschi,
A. Schwehn,
C. Schröder,
M. Weinmann,
and F. Werner
contributed both to the design and implementation of the
presented concepts as well as the contents of the research
paper. 
M. Lienkamp, B. Lohmann, and J. Betz made an essential contribution to
the concept of the research project. They revised the paper critically for important intellectual content. 
M. Lienkamp gives final approval for the version to be published and
agrees to all aspects of the work. As a guarantor, he accepts
responsibility for the overall integrity of the paper.

\paragraph*{Competing interests:}
There are no competing interests to declare.
\clearpage
\paragraph*{Data and materials availability:}
In order to accelerate the development of autonomous driving in general, we provide large portions of our software stack as open-source software.

\begin{itemize}
    \item Multibody Vehicle Dynamics Simulation:\\
    \url{https://github.com/TUMFTM/Open-Car-Dynamics}
    \item 3D Vehicle State Estimation:\\
    \url{https://github.com/TUMFTM/3DVehicleDynamicsStateEstimation}
    \item Georeferencing of Point Cloud Maps:\\
    \url{https://github.com/TUMFTM/FlexCloud}
    \item Object Tracking:\\
    \url{https://github.com/TUMFTM/FusionTracking}
    \item 3D Raceline Planning:\\
    \url{https://github.com/TUMRT/online_3D_racing_line_planning}
    \item 3D Sampling-based Local Trajectory Planning:\\
    \url{https://github.com/TUMRT/sampling_based_3D_local_planning}
    \item Longitudinal Vehicle Dynamics Control:\\
    \url{https://github.com/TUMFTM/TAM__long_acc_control}
    \item Trajectory Tracking Controller:\\
    \url{https://github.com/TUMFTM/mod_vehicle_dynamics_control}
    \item GGGV Diagram Generation:\\
    \url{https://github.com/TUM-AVS/GGGVDiagrams}
    \item Time Series Logging (TSL) Library:\\
    \url{https://github.com/TUMFTM/tsl}
    \item TAM Param Management:\\
    \url{https://github.com/TUMFTM/TAM__param_management}
    \item TAM Common Functionality:\\
    \url{https://github.com/TUMFTM/TAM__common}
    \item TAM Messages:\\
    \url{https://github.com/TUMFTM/TAM__msgs}
\end{itemize}

\end{document}